\documentclass{article}

\usepackage[final,nonatbib]{nips_2016}

\usepackage[utf8]{inputenc} 
\usepackage[T1]{fontenc}    
\usepackage{url}            
\usepackage{booktabs}       
\usepackage{amsfonts}       
\usepackage{nicefrac}       
\usepackage{microtype}      
\usepackage{graphicx}
\PassOptionsToPackage{square,numbers}{natbib}

\newcommand{\yy}[1]{\textit{TODO: {#1}}}

\newcommand{\cd}[1]{{\small \texttt{#1}}}

\newcommand{\mysubsection}[1]{\textbf{{#1}.}}

\newif\iflong
\longfalse

\newcommand{\longonly}[1]{\iflong{#1}\fi}

\newcommand{\trm}[1]{\textit{#1}}
\newcommand{\vek}[1]{\textbf{#1}}

\newcommand{\M}{\textbf{M}}
\newcommand{\exs}{{\cal E}x}
\newcommand{\model}{\textit{Model}}
\newcommand{\C}{{\cal{C}}}
\newcommand{\T}{{\cal{T}}}
\newcommand{\DB}{{\cal{DB}}}
\newcommand{\onenorm}[1]{{|\!|{#1}|\!|_1}}

\newcommand{\indicate}[1]{|\![{#1}]\!|}
\newenvironment{alg}{\begin{minipage}[t]{\textwidth}\begin{tabbing}12\=12\=12\=12\=12\=12\=12\=12\=\kill}{\end{tabbing}\end{minipage}}

\title{TensorLog: A Differentiable Deductive Database}

\author{
  William W. Cohen\\
  Department of Machine Learning\\
  Carnegie Mellon University\\
  Pittsburgh, PA 15213 \\
  \texttt{wcohen@cs.cmu.edu} \\
}

\begin{document}

\maketitle

\begin{abstract}
Large knowledge bases (KBs) are useful in many tasks, but it is
unclear how to integrate this sort of knowledge into ``deep''
gradient-based learning systems.  To address this problem, we describe
a probabilistic deductive database, called TensorLog, in which
reasoning uses a differentiable process.  In TensorLog, each clause in
a logical theory is first converted into certain type of factor graph.
Then, for each type of query to the factor graph, the message-passing
steps required to perform belief propagation (BP) are ``unrolled''
into a function, which is differentiable.  We show that these
functions can be composed recursively to perform inference in
non-trivial logical theories containing multiple interrelated clauses
and predicates.  Both compilation and inference in TensorLog are
efficient: compilation is linear in theory size and proof depth, and
inference is linear in database size and the number of message-passing
steps used in BP.  We also present experimental results with TensorLog
and discuss its relationship to other first-order probabilistic
logics.
\end{abstract}

\section{Introduction}

Large knowledge bases (KBs) have proven useful in many tasks, but it
is unclear how to integrate this sort of knowledge into ``deep''
gradient-based learning systems.  Motivated by this, we describe a
probabilistic deductive database (PrDDB) system in which reasoning is
performed by a differentiable process.  In addition to enabling novel
gradient-based learning algorithms for PrDDBs, this approach could
potentially enable tight integration of logical reasoning into deep
learners (or conversely, of deep learning into reasoning systems.

In a traditional deductive database (DDB), a database $\DB$ with a
theory $\T$ together define a set of facts $f_1,\ldots,f_n$ which can
be derived by accessing the database and reasoning using $\T$.  As an
example, Figure~\ref{fig:ddb} contains a small theory and an
associated database.  End users can test to see if a fact $f$ is
derivable, or retrieve all derivable facts that match some query:
e.g., one could test if $f=\cd{uncle(joe,bob)}$ is derivable in the
sample database, or find all values of $Y$ such that \cd{uncle(joe,Y)}
holds.  A probabilistic DDB is a ``soft'' extension of a DDB, where
derived facts $f$ have a numeric confidence, typically based on
augmenting $\DB$ with a set of parameters $\Theta$.  In many existing
PrDDB models computation of confidences is computationally expensive,
and often not be conducive to learning the parametersb $\Theta$.

Here we describe a probabilistic deductive database called TensorLog
in which reasoning uses a differentiable process.  In TensorLog, each
clause in a logical theory is first converted into certain type of
factor graph, in which each logical variable appearing in the clause
is associated with a random variable in the factor graph, and each
literal is associated with a factor (as shown in
Figure~\ref{fig:factors}).  Then, for each type of query to the factor
graph, the message-passing steps required to perform BP are
``unrolled'' into a function, which is differentiable.  Each function
will answer queries for a particular combination of evidence variables
and query variables in the factor graph, which in turn corresponds to
logical queries in a particular syntactic form.  We also show how
these functions can be composed recursively to perform inference in
non-trivial logical theories containing multiple interrelated clauses
and predicates.

In TensorLog, compilation is linear in theory size and proof depth,
and inference is linear in database size and the number of
message-passing steps used in BP.  Most importantly, \emph{inference
  is also differentiable}, enabling gradient-based parameter
learning. Formally, we can show that TensorLog subsumes some prior
probabilistic logic programming models, including several variants of
stochastic logic programs (SLPs)
\cite{DBLP:journals/ml/Cussens01,wang2013programming}, and
approximates others
\cite{poole1997independent,fierens2016}.

Below, we first present background material, then introduce our main
results for differentiable inference, We then discuss related work, in
particular the relationship between TensorLog and existing
probabilistic logics, present experimental results, and conclude.

\section{Background: Deductive and Probabilistic DBs}

\begin{figure}
\begin{small}
\begin{center}
\begin{tabular}[t]{l}
1. \cd{uncle(X,Y):-child(X,W),brother(W,Y).}\\ 2. \cd{uncle(X,Y):-aunt(X,W),husband(W,Y).}\\ 3. \cd{status(X,tired):-child(W,X),infant(W).}\\
\end{tabular}\begin{tabular}[t]{ll}
\cd{child(liam,eve)},0.99   & \cd{infant(liam)},0.7\\
\cd{child(dave,eve)},0.99   & \cd{infant(dave)},0.1\\
\cd{child(liam,bob)},0.75   & \cd{aunt(joe,eve)},0.9\\
\cd{husband(eve,bob)},0.9   & \cd{brother(eve,chip)},0.9\\
\end{tabular}
\end{center}
\end{small}
\caption{\small An example database and theory.  Uppercase symbols are
  universally quantified variables, and so clause 3 should be read as
  a logical implication: for all database constants $c_X$ and $c_W$,
  if \cd{child($c_X$,$c_W$)} and \cd{infant($c_W$)} can be proved,
  then \cd{status($c_X$,tired)} can also be proved.}\label{fig:ddb}
\end{figure}

To begin, we review the definition for an ordinary DDB, an example of
which is in Figure~\ref{fig:ddb}.  A \trm{database}, $\DB$, is a set
$\{f_1,\ldots,f_N\}$ of ground facts.  We focus here on DB relations
which are unary or binary (e.g., from a ``knowledge graph''), hence,
facts will be written as $p(a,b)$ or $q(c)$ where $p$ and $q$ are
\trm{predicate symbols}, and $a,b,c$ are constants from a fixed domain
$\C$.  A theory, $\T$, is a set of function-free Horn clauses. Clauses
are written $A\cd{:-}B_1,\ldots,B_k$, where $A$ is called the
\trm{head} of the clause, $B_1,\ldots,B_k$ is the \trm{body}, and $A$
and the $B_i$'s are called \trm{literals}. Literals must be of the
form $q(X)$, $p(X,Y)$, $p(c,Y)$, or $p(X,c)$, where $X$ and $Y$ are
logical variables, and $c$ is a database constant.

Clauses can be understood as logical implications.  Let $\sigma$ be a
\trm{substitution}, i.e., a mapping from logical variables to
constants in $\C$, and let $\sigma(L)$ be the result of replacing all
logical variables $X$ in the literal $L$ with $\sigma(X)$.  A set of
tuples $S$ is \trm{deductively closed} with respect to the clause
$A\leftarrow{}B_1,\ldots,B_k$ iff for all substitutions $\sigma$,
either $\sigma(A) \in {}S$ or $\exists B_i:\sigma(B_i)\not\in{}S$.
For example, if $S$ contains the facts of Figure~\ref{fig:ddb}, $S$ is
not deductively closed with respect to the clause 1 unless it also
contains \cd{uncle(chip,liam)} and \cd{uncle(chip,dave)}.  The
\trm{least model} for a pair $\DB,\T$, written $\model(\DB,\T)$, is
the smallest superset of $\DB$ that is deductively closed with respect
to every clause in $\T$.  In the usual DDB semantics, a ground fact
$f$ is considered ``true'' iff $f\in\model(\DB,T)$.

To introduce ``soft'' computation into this model, we add a parameter
vector $\Theta$ which associates each fact $f\in\DB$ with a positive
scalar $\theta_f$ (as shown in the example).  The semantics of this
parameter vary in different PrDDB models, but $\Theta$ will always
define a distribution $\Pr(f|\T,\DB,\Theta)$ over the facts in
$\model(\T,\DB)$.

\section{Differentiable soft reasoning}

\mysubsection{Numeric encoding of PrDDB's and queries}
We will implement reasoning in PrDDB's by defining a series of numeric
functions, each of finds answers to a particular family of queries.
It will be convenient to encode the database numerically. We will
assume all constants have been mapped to integers.  For a constant
$c\in\C$, we define $\vek{u}_c$ to be a one-hot row-vector
representation for $c$, i.e., a row vector of dimension $|\C$| where
$\vek{u}[c]=1$ and $\vek{u}[c']=0$ for $c'\not=C$.  We can also
represent a binary predicate $p$ by a sparse matrix $\M_p$, where
$\M_p[a,b]=\theta_{p(a,b)}$ if $p(a,b)\in\DB$, and a unary predicate
$q$ as an analogous row vector $\vek{v}_q$.  Note that $\M_p$ encodes
information not only about the database facts in predicate $p$, but
also about their parameter values.

PrDDB's are commonly used test to see if a fact $f$ is derivable, or
retrieve all derivable facts that match some query: e.g., one could
test if $f=\cd{uncle(joe,bob)}$ is derivable in the sample database,
or find all values of $Y$ such that \cd{uncle(joe,Y)} holds.  We focus
here on the latter type of query, which we call an {argument-retrieval
  query}.  An \trm{argument-retrieval query} $Q$ is of the form
$p(c,Y)$ or $p(Y,c)$: we say that $p(c,Y)$ has an \trm{input-output
  mode} of \cd{in,out} and $p(Y,c)$ has an input-output mode of
\cd{out,in}. For the sake of brevity, below we will assume below the
mode \cd{in,out} when possible, and abbreviate the two modes as
\cd{io} and \cd{io}.

The \trm{response} to a query $p(c,Y)$ is a distribution over possible
substitutions for $Y$, encoded as a vector $\vek{v}_{Y}$ such that for
all constants $d\in\C$, $\vek{v}_{Y}[d] = \Pr(p(c,d)|\T,\DB,\Theta)$.
Alternatively (since often we care only about the relative scores of
the possible answers), the system might instead return a conditional
probability vector $\vek{v}_{Y|c}$: if $U_{p(c,Y)}$ is the set
of facts $f$ that ``match'' $p(c,Y)$, then $\vek{v}_{Y|c}[d] =
\Pr(f=p(c,d)|f\in{}U_{p(c,Y)},\T,\DB,\Theta)$.

\longonly{Although here we only consider single-literal queries, more
  complex queries can be answered by extending the theory: e.g., to
  find $\{ Y: \cd{uncle(joe,X),husband(X,$Y$)}\}$, we could add the
  clause \cd{q1(J,Y):-uncle(joe,X),husband(X,Y)} to the theory and
  find the answer to \cd{q1(joe,Y)}.}

Since the ultimate goal of our reasoning system is to correctly answer
queries using functions, we also introduce a notation for functions
that answer particular types of queries: in particular, for a
predicate symbol $f^p_\cd{io}$ denotes a \trm{query response function}
for all queries with predicate $p$ and mode \cd{io}, i.e., queries of
the form $p(c,Y)$, when given a one-hot encoding of $c$, $f^p_\cd{io}$
returns the appropriate conditional probability vector:
\begin{equation} \label{eq:def-f}
 f^p_\cd{io}(\vek{u}_c) \equiv \vek{v}_{Y|X} 
    \mbox{~where~} \forall d\in C: \vek{v}_{Y|c}[d] =
             \Pr(f=p(c,d)|f\in{}U_{p(c,Y)},\T,\DB,\Theta)
\end{equation}
and similarly for $f^p_\cd{oi}$.

\mysubsection{Syntactic restrictions}
Algorithmically it will be convenient to constrain the use of
constants in clauses.  We introduce a special DB predicate
\cd{assign}, which will be used only in literals of the form
\cd{assign($W$,$c$)}, which in turn will be treated as literals for a
special unary predicate \cd{assign\_$c$}.  Without loss of generality,
we can now assume that constants only appear in \cd{assign} literals.
For instance, the clause 3 of Figure~\ref{fig:ddb} would be rewritten
as
\begin{equation} \label{eq:tired}
\cd{status(X,T):-assign\_tired(T),child(X,W),infant(W).}
\end{equation}
We will also introduce another special DB predicate \cd{any}, where
\cd{any($a,b$)} is conceptually true for any pair of constants $a,b$;
however, as we show below, the matrix $\M_\cd{any}$ need not be
explicitly stored.  We also constrain clause heads to contain distinct
variables which all appear also in the body.

\begin{figure}
\centerline{\includegraphics[width=0.8\linewidth]{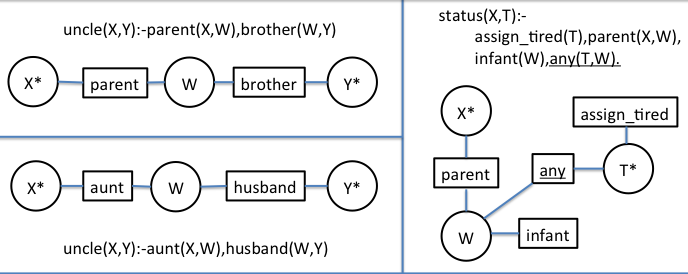}}
\caption{\small Examples of factor graphs for the example
  theory.} \label{fig:factors}
\end{figure}

\mysubsection{A factor graph for a one-clause program} We will start
by considering a highly restricted class of theories $\T$, namely
programs containing only one non-recursive clause $r$ that obeys the
restrictions above.  We build a factor graph $G_r$ for $r$ as follows:
for each logical variable $W$ in the body, there is a random variable
$W$; and for every literal $q(W_i,W_j)$ in the body of the clause,
there is a factor with potentials $\M_q$ linking variables $W_i$ and
$W_j$.  Finally, if the factor graph is disconnected, we add \cd{any}
factors between the components until it is connected.
Figure~\ref{fig:factors} gives examples.  The variables appearing in
the clause's head are starred.

We now argue that $G_r$ imposes a valid distribution
$\Pr(f|\T,\DB,\Theta)$ over facts in $\model(\T,\DB)$.  In $G_r$ the
variables are multinomials over $\C$, the factors represent predicates
and the graph $G_r$ represents a distribution of \emph{possible
  bindings to the logical variables in the clause $f$}, i.e., to
possible substitutions $\sigma$. Let $W_1,\ldots,W_m$ be the variables
of $G_r$, and for each factor/edge $e$ let $p_e(W_{i_e},W_{j_e})$ be
the literal associated with it. In the distribution defined by $G_r$
\[
 \Pr_{G_r}(W_1=c_1,\ldots,W_m=c_m) = \frac{1}{Z}
 \prod_{(c_i,c_j)\in\textit{edges $e$}} \phi_e(c_i,c_j) =
 \prod_{(c_i,c_j)\in\textit{edges $e$}} \theta_{p_e(c_{i_e},c_{j_e})}
\]
Recall $\forall f,\theta_f>0$, so if $\Pr(W_1=c_1,\ldots,W_m=c_m)>0$
then for each edge $p_e(c_{i_e},c_{j_e}) \in \DB$, and hence the
substitution $\sigma=\{W_1=c_1,\ldots,W_m=c_m\}$ makes the body of
clause $r$ true.  The converse is also clearly true: so $G_r$ defines
a distribution over exactly those substitutions $\sigma$ that make the
the body of $r$ true.  

\begin{figure}
\begin{center}
\begin{small}
\begin{alg}
\textbf{define} compileMessage($L \rightarrow X$):\\
\> assume wolg that $L=q(X)$ or $L=p(X_i,X_o)$\\
\> generate a new variable name  $\vek{v}_{L,X}$  \\
\> \textbf{if} $L=q(X)$ \textbf{then}\\
\> \>  emitOperation( $\vek{v}_{L,X} = \vek{v}_q$)\\
\> \textbf{else if} $X$ is the output variable $X_o$ of $L$ \textbf{then}\\
\> \> $\vek{v}_i =$ compileMessage($X_i \rightarrow L$)\\
\> \> emitOperation( $\vek{v}_{L,X} = \vek{v}_i \cdot \M_p$ )\\
\> \textbf{else if} $X$ is the input variable $X_i$ of $L$ \textbf{then}\\
\> \> $\vek{v}_o =$ compileMessage($X_i \rightarrow L$) \\%
\> \> emitOperation( $\vek{v}_{L,X} = \vek{v}_o \cdot \M_p^T$ ) \\ 
\> \textbf{return} $\vek{v}_{L,X}$\\
\end{alg}~~~~\begin{alg}
\textbf{define} compileMessage($X \rightarrow L$):  \\
\> \textbf{if} $X$ is the input variable $X$ \textbf{then}\\
\> \> \textbf{return} $\vek{u}_c$, the input\\
\> \textbf{else}\\
\> \> generate a new variable name $\vek{v}_X$\\
\> \> assume $L_1,L_2,\ldots,L_k$ are the \\
\> \> ~~ neighbors of $X$ excluding $L$ \\
\> \> \textbf{for} $i=1,\ldots,k$ \textbf{do}\\
\> \> \> $\vek{v}_i =$ compileMessage($L_i \rightarrow X$)\\
\> \> emitOperation($\vek{v}_X = \vek{v}_1 \circ \cdots \circ \vek{v}_k$) \\
\> \> \textbf{return} $\vek{v}_X$\\
\end{alg}
\end{small}
\end{center}
\caption{\small Algorithm for unrolling belief propagation on a
  polytree into a sequence of message-computation operations. Notes:
  (1) if $L=p(X_o,X_i)$ then replace $\M_p$ with $\M_p^T$ (the
  transpose). (2) Here $\vek{v}_1 \circ \vek{v}_2$ denotes the
  Hadamard (component-wise) product, and if $k=0$ an all-ones vector
  is returned.} \label{fig:alg}
\end{figure}

BP over $G_r$ can now be used to compute the conditional vectors
$f^p_\cd{io}(\vek{u}_c)$ and $f^p_\cd{oi}(\vek{u}_c)$.  For example to
compute $f^p_\cd{io}(\vek{u}_c)$ for clause 1, we would set the
message for the evidence variable $X$ to $\vek{u}_c$, run BP, and read
out as the value of $f$ the marginal distribution for $Y$.  

However, we would like to do more: we would like to compute an
explicit, differentiable, query response function, which computes
$f^p_\cd{io}(\vek{u}_c)$.  To do this we ``unroll'' the
message-passing steps into a series of operations, following
\cite{gormley_approximation-aware_2015}.

For completeness, we include in Figure~\ref{fig:alg} a sketch of the
algorithm used in the current implementation of TensorLog, which makes
the (strong) assumption that $G_r$ is a tree.
In the code, we found it convenient to extend the notion of
input-output modes for a query, a variable $X$ appearing in a literal
$L=p(X,Y)$ in a clause body is an \trm{nominal input} if it appears in
the input position of the head, or any literal to the left of $L$ in
the body, and is an \trm{nomimal output} otherwise.  In Prolog a
convention is that nominal inputs appear as the first argument of a
predicate, and in TensorLog, if the user respects this convention,
then ``forward'' message-passing steps use $M_p$ rather than $M_p^T$w
(reducing the cost of transposing large $\DB$-derived matrices, since
our message-passing schedule tries to maximize forward messages.)  The
code contains two mutually recursive routines, and is invoked by
requesting a message from the output variable to a fictional output
literal. The result will be to emit a series of operations, and return
the name of a register that contains the unnormalized conditional
probability vector for the output variable: e.g., for the sample clauses
the functions returned are:

\begin{small}
\begin{tabular}{lll} 
 r1 & $g^{r1}_\cd{io}(\vec{u}_c) = \{\!\!\!\!\!$ &
            $\vek{v}_{1,W} = \vek{u}_c \M_\cd{parent}$; $\vek{v}_W = \vek{v}_{1,W}$; $\vek{v}_{2,Y} = \vek{v}_W \M_\cd{brother}$; $\vek{v}_Y = \vek{v}_{2,Y}$;
            ~ \textbf{return} $\vek{v}_{Y} \}$ \\
 r2 & $g^{r2}_\cd{io}(\vec{u}_c) = \{\!\!\!\!\!$ &
            $\vek{v}_{1,W} = \vek{u}_c \M_\cd{aunt}$; $\vek{v}_W = \vek{v}_{1,W}$; $\vek{v}_{2,Y} = \vek{v}_W \M_\cd{husband}$; $\vek{v}_Y = \vek{v}_{2,Y}$;
            ~ \textbf{return} $\vek{v}_{Y} \}$ \\
 r3 & $g^{r3}_\cd{io}(\vec{u}_c) = \{\!\!\!\!\!$ &
            $\vek{v}_{2,W} = \vek{u}_c \M_\cd{parent}$; $\vek{v}_{3,W} = \vek{v}_\cd{infant}$; $\vek{W} = \vek{v}_{2,W} \circ \vek{v}_{3,W}$;\\
       & &  $\vek{v}_{1,T} = \vek{v}_\cd{assign\_tired}$; $\vek{v}_{4,T} = \vek{v}_{W} \M_\cd{any}$; $\vek{T} = \vek{v}_{1,T} \circ \vek{v}_{4,T}$; 
            ~ \textbf{return} $\vek{v}_{T} \}$ \\
\end{tabular}
\end{small}

Here we use $g^r_\cd{io}(\vec{u}_c)$ for the unnormalized version of
the query response function build from $G_r$, i.e.,
\[ f^p_\cd{io}(\vec{u}_c) \equiv  g^r_\cd{io}(\vec{u}_c)/\onenorm{g^r_\cd{io}(\vec{u}_c)} 
\]
where $r$ is the one-clause theory defining $p$.  

\mysubsection{Sets of factor graphs for multi-clause programs} \label{sec:multi-clause}
We now extend this idea to theories with many clauses.  We first note
that if there are several clauses with the same predicate symbol in
the head, we simply sum the unnormalized query response functions:
e.g., for the predicate cd{uncle}, defined by rules $r_1$ and $r_2$,
we can define
\[ g^\cd{uncle}_\cd{io} = g^{r1}_\cd{io} +  g^{r2}_\cd{io}
\]
and then re-normalize.  This is equivalent to building a new factor
graph $G$, which would be approximately $\cup_i G_{ri}$, together
global input and output variables, and a factor that constrains the
input variables of the $G_{ri}$'s to be equal, and a factor that
constrains the output variable of $G$ to be the sum of the outputs of
the $G_{ri}$'s.

A more complex situation is when the clauses for one predicate, $p$,
use a second theory predicate $q$, in their body: for example, this
would be the case if \cd{aunt} was also defined in the theory, rather
than the database.  For a theory with no recursion, we can replace the
message-passing operations $\vek{v}_Y = \vek{v}_X \M_q$ with the
function call $\vek{v}_Y = g^q_\cd{io}(\vek{v}_X)$, and likewise the
operation $\vek{v}_Y = \vek{v}_X \M_q^T$ with the function call
$\vek{v}_Y = g^q_\cd{oi}(\vek{v}_X)$.  It can be shown that this is
equivalent to taking the factor graph for $q$ and ``splicing'' it into
the graph for $p$.

It is also possible to allow function calls to recurse to a fixed
maximum depth: we must simply add some sort of extra argument that
tracks depth to the recursively-invoked $g^q$ functions, and make sure
that $g^p$ returns an all-zeros vector (indicating no more proofs can
be found) when the depth bound is exceeded.  Currently this is
implemented by marking learned functions $g$ with the predicate $q$, a
mode, and a depth argument $d$, and ensuring that function calls
inside $g^p_{\cd{io},d}$ to $q$ always call the next-deeper version of
the function for $q$, e.g., $g^q_{\cd{io},d+1}$.

\mysubsection{Uncertain inference rules} \label{sec:soft-rules} Notice
that $\Theta$ associates confidences with \emph{facts} in the
databases, not with \emph{clauses} in the theory.  To attach a
probability to a clause, a standard trick is to introduce a special
clause-specific fact, and add it to the clause body
\cite{poole1997independent}.  For example, a soft version of clause 3
could be re-written as
\[
\cd{status(X,tired):-assign(RuleId,c3),weighted(RuleId),child(W,X),infant(W)}
\]
where the (parameterized) fact \cd{weighted(c3)} appears in $\DB$, and
the constant \cd{c3} appears nowhere else in $\T$.  TensorLog supports
some special syntax to make it easy to build rules with associated
weights: for instance, \cd{status(X,tired)} :- \cd{assign(C3,c3),
  weighted(C3), child(W,X), infant(W)} can be written simply as
\cd{status(X,tired)} :- \cd{child(W,X), infant(W) \{c3\}}.  \longonly{
  One can also attach a computed set of features to a rule in order to
  weight it, as in ProPPR: e.g., one can write \cd{status(X,tired)} :-
  \cd{\{all(A):child(W,X),age(W,A)\}}, which indicates that the all
  the ages of the children of $X$ should be used as features to
  determine if the rule succeeds, and it will be expanded to
  \cd{status(X,tired)} :- \cd{child(W,X),age(W,A),weighted(A)}.  These
  features are used in the experiments below, where we compare to
  ProPPR.}

\mysubsection{Discussion}
Collectively, the computation performed by TensorLog's functions are
equivalent to computing a set of marginals over a particular factor
graph $G$: specifically $G$ would be formed by using the construction
for multiple clauses with the same head (described above), and then
splicing in the factor graphs of subpredicates.  The unnormalized
messages over this graph, and their functional equivalent, can be
viewed implementing a first-order version of \trm{weighted model
  counting}, a well-studied problem in satisfiability.

Computationally, the algorithm we describe is quite
efficient. Assuming the matrices $\M_p$ exist, the additional memory
needed for the factor-graph $G_r$ is linear in the size of the clause
$r$, and hence the compilation to response functions is linear in the
theory size and the number of steps of BP.  For theories where every
$G_r$ is a tree, the number of message-passing steps is also linear.
Message size is (by design) limited to $|\C|$, and is often smaller
due to sparsity.

The current implementation of TensorLog includes many restrictions
that could be relaxed: e.g., predicates must be unary or binary, only
queries of the types discussed here are allowed, and every factor
graph $G_r$ must be a tree. 
Matrix operations are implemented in the
scipy sparse-matrix package, and the ``unrolling'' code performs a
number of optimizations to the sequence in-line: one important one is
to use the fact that \( \vek{v}_X \circ (\vek{v}_Y \M_\cd{any}) =
\vek{v}_X \onenorm{\vek{v}_Y} \) to avoid explicitly building
$\M_\cd{any}$.  

\section{Related Work}

\mysubsection{Hybrid logical/neural systems} There is a long tradition
of embedding logical expressions in neural networks for the purpose of
learning, but generally this is done indirectly, by conversion of the
logic to a boolean formula, rather than developing a differentiable
theorem-proving mechanism, as considered here.  Embedding logic may
lead to a useful architecture \cite{TowellAAAI90} or regularizer
\cite{riedelinjecting2015}.

Recently \cite{rocktaschel2016learning} have proposed a differentiable
theorem prover, in which a proof for an example is unrolled into a
network.  Their system includes representation-learning as a
component, as well as a template-instantiation approach (similar to
\cite{wang2014structure}), allowing structure learning as well.
However, published experiments with the system been limited to very
small datasets.  Another recent paper
\cite{DBLP:journals/corr/AndreasRDK16} describes a system in which
non-logical but compositionally defined expressions are converted to
neural components for question-answering tasks.

\mysubsection{Explicitly grounded probabilistic first-order languages}
Many first-order probabilistic models are implemented by
``grounding'', i.e., conversion to a more traditional
representation.\footnote{For a survey of such models see
  \cite{kimmig2015lifted}.}  For example, Markov logic networks (MLNs)
are a widely-used probabilistic first-order model
\cite{RichardsonMLJ2006} in which a Bernoulli random variable is
associated with each \emph{potential} ground database fact (e.g., in
the binary-predicate case, there would be a random variable for each
possible $p(a,b)$ where $a$ and $b$ are any facts in the database and
$p$ is any binary predicate) and each ground instance of a clause is a
factor.  The Markov field built by an MLN is hence of size $O(|\C|^2)$
for binary predicates, which is much larger than the factor graphs
used by TensorLog, which are of size linear in the size of the theory.
In our experiments we compare to ProPPR, which has been elsewhere
compared extensively to MLNs.

Inference on the Markov field can also be expensive, which motivated
the development of probabilistic similarity logic (PSL),
\cite{brocheler2012probabilistic} a MLN variant which uses a more
tractible hinge loss, as well as lifted relational neural networks
\cite{DBLP:journals/corr/SourekAZK15}, a recent model which grounds
first-order theories to a neural network.  However, any grounded model
for a first-order theory can be very large, limiting the scalability
of such techniques.

\mysubsection{Probabilistic deductive databases and tuple independence}
\longonly{
\begin{table} \label{tab:queries}
  \caption{\small Explanations for some sample queries.  For the
    queries marked with a $*$, we use a modified theory where the
    third clause was replaced with these three clauses:
    \cd{status(X,tired):-caresFor(X,W),infant(W)};
    \cd{caresFor(T,U):-child(T,U)}; and
    \cd{caresFor(T,U):-child(T,W),husband(W,U)}.  }
\begin{small}
\newcommand{\dex}{$\exs(f)=\{\!\!\!\!\!$}
\begin{tabular}{lrl}
\multicolumn{1}{c}{Fact $f$} & & \multicolumn{1}{c}{Explanations for $f$}  \\ \hline
 \cd{status(eve,tired)}  & \dex         & $\{\cd{(child(eve,liam)},\cd{infant(liam)}\},$ \\ 
                         &              & $\{\cd{(child(eve,dave)},\cd{infant(dave)}\} ~~~\}$ \\ 
 \cd{status(bob,tired)}  & \dex         & $\{\cd{(child(bob,liam)},\cd{infant(liam)}\} ~~~\}$ \\ \hline\hline
 \cd{status(eve,tired)}$^*$  
                         & \dex         & $\{\cd{(child(eve,liam)},\cd{infant(liam)}\}$, \\ 
                         &              & $\{\cd{(child(eve,dave)},\cd{infant(dave)}\} ~~~\}$ \\ 
 \cd{status(bob,tired)}$^*$  
                         & \dex         & $\{\cd{(child(bob,liam)},\cd{infant(liam)}\}$, \\ 
                         &              & $\{\cd{(child(eve,liam)},\cd{husband(eve,bob)},\cd{infant(liam)}\}$, \\ 
                         &              & $\{\cd{(child(eve,dave)},\cd{husband(eve,bob)},\cd{infant(dave)}\} ~~~\}$ \\ \hline
\end{tabular}
\end{small}
\end{table}}

TensorLog is superficially similar to the \trm{tuple independence}
model for PrDDB's \cite{suciu2011probabilistic}, which use $\Theta$ to
define a distribution, $\Pr(I|\DB,\Theta)$, over ``hard'' databases
(aka \trm{interpretations}) $I$.  In particular, to generate $I$, each
fact $f\in\DB$ sampled by independent coin tosses, i.e.,
\( 
\Pr_\cd{TupInd}(I|\DB,\Theta) \equiv \prod_{t \in I} \theta_t \cdot
\prod_{t \in \DB-I} (1-\theta_t) 
\). The
probability of a derived fact $f$ is defined as follows, where
$\indicate{\cdot}$ is a zero-one indicator function:
\begin{equation} \label{eq:ddb-semantics}
 \Pr_\cd{TupInd}(f|\T,\DB,\Theta) \equiv \sum_{I} \indicate{f\in\model(I,\T)} \cdot \Pr(I|\DB,\Theta)
\end{equation}

There is a large literature (for surveys, see
\cite{suciu2011probabilistic,de2008probabilistic}) on approaches to
tractibly estimating Eq~\ref{eq:ddb-semantics}, which naively requires
marginalizing over all $2^{|\DB|}$ interpretations.  One approach, taken
by the ProbLog system \cite{fierens2016}, relies on the notion of an
\trm{explanation}.  An \trm{explanation} $E$ for $f$ is a
\emph{minimal} interpretation that supports $f$: i.e.,
$f\in{}\model(\T,E)$ but $f\not\in{}\model(\T,E')$ for all
$E'\subset{}E$.  It is easy to show that if $E'' \supset{} E$ then
$f\in{}\model(\T,E'')$; hence, the set $\exs(f)$ of all explanations
for $f$ is a more concise representation of the interpretations that
support $f$.  \longonly{Some sample explanations are shown in
  Table~\ref{tab:queries}.}

Under the tuple independence model\longonly{ of
  Eq~\ref{eq:tuple-indep}}, the marginal probability of drawing some
interpretation $I\supseteq{}E$ is simply
\[ \sum_{I \supseteq E} \prod_{f'\in{}I} \theta_{f'} \prod_{f'\in{}\DB-I} (1-\theta_{f'}) = \prod_{f'\in{}E} \theta_{f'}
\]
while in TensorLog, 
\[ \Pr_\cd{TenLog}(f) = \frac{1}{Z} g_\cd{TenLog}(f), \mbox{~where~} g_\cd{TenLog}(f) = \sum_{E\in{}\exs(f)} \prod_{f'\in{}E} \theta_{f'} 
\]
So TensorLog's score for a single-explanation fact is the same as under $\Pr_\cd{TupInd}$, but more generally only
approximates Eq~\ref{eq:ddb-semantics}, since 
\begin{eqnarray*}
 \Pr_\cd{TupInd}(f) & =  & \sum_{I} \indicate{f\in\model(I,\T)} \cdot \Pr(I) = \sum_{I:f\in\model(I,\T)} \prod_{f'\in{}I} \theta_{f'}\prod_{f'\in{}\DB-I} (1-\theta_{f'}) \\
       & = & \sum_{E\in{}\exs(I)} \sum_{I\supseteq E} \prod_{f'\in{}I} \theta_{f'} \not= \sum_{E\in\exs(f)} \prod_{f'\in{}E} \theta_{f'} = g_\cd{TenLog}(f)
\end{eqnarray*}
the inequality occurring because TensorLog overcounts interpretations
$I$ that are supersets of more than one explanation.  

This approximation step is important to TensorLog's efficiency,
however. Exact computation of probabilities in the tuple independence
model are \#P hard to compute \cite{fierens2016} \trm{in
  the size of the set of explanations}, which as noted, can itself be
exponentially large.  A number of methods have been developed for
approximating this computation, or performing it as efficiently as can
be done---for example, by grounding to a boolean formula and
converting to a decision-tree like format that facilitates counting
\cite{suciu2011probabilistic}.  Below we experimentally compare
inference times to ProbLog2, one system which adopts these semantics.

\mysubsection{Stochastic logic programs and ProPPR} \label{sec:SLPs}
TensorLog is more closely related to stochastic logic programs (SLPs)
\cite{DBLP:journals/ml/Cussens01}. In an SLP, a probabilistic process
is associated with a top-down theorem-prover: i.e., each clause $r$
used in a derivation has an assocated probability $\theta_{r}$.  Let
$N(r,E)$ be the number of times $r$ was used in deriving the
explanation $E$: then in SLPs,
\( \Pr_\cd{SLP}(f) = \frac{1}{Z} \sum_{E\in\exs(f)} \prod_r \theta_r^{N(r,E)}
\). 
The same probability distribution can be generated by TensorLog if (1)
for each rule $r$, the body of $r$ is prefixed with the literals
\cd{assign(RuleId,$r$),weighted(RuleId)}, where $r$ is a unique
identifier for the rule and (2) $\Theta$ is constructed so that
$\theta_f=1$ for ordinary database facts $f$, and
$\theta_\cd{weighted(r)}=\theta'_\cd{r}$, where $\Theta'$ is the
parameters for a SLP.

SLPs can be \trm{normalized} or \trm{unnormalized}; in normalized
SLPs, $\Theta$ is defined so for each set of clauses $S_p$ of clauses
with the same predicate symbol $p$ in the head, $\sum_{r\in{}S_p}
\theta_r=1$.  TensorLog can represent both normalized and unnormalized
SLPs (although clearly learning must be appropriately constrained to
learn parameters for normalized SLPs.)  Normalized SLPs generalize
probabilistic context-free grammars, and unnormalized SLPs can express
Bayesian networks or Markov random fields
\cite{DBLP:journals/ml/Cussens01}.

ProPPR \cite{wang2013programming} is a variant of SLPs in which (1)
the stochastic proof-generation process is augmented with a reset, and
(2) the transitional probabilities are based on a normalized
soft-thresholded linear weighting of features.  The first extension to
SLPs can be easily modeled in TensorLog, but the second cannot: the
equivalent of ProPPR's clause-specific features can be incorporated,
but they are globally normalized, not locally normalized as in ProPPR.

ProPPR also includes an approximate grounding procedure which
generates networks of size linear in $m$, $\alpha^{-1}$,
$\epsilon^{-1}$, and where $m$ is the number of training examples,
$\alpha$ is the reset parameter, $\it{deg}_{it max}$ is the maximum
degree of the proof graph, and $\epsilon$ is the pointwise error of
the approximation.  Asymptotic analysis suggests that ProPPR should be
faster for very large database and small numbers of training examples
(assuming moderate values of $\epsilon$ and $\alpha$ are feasible to
use), but that TensorLog should be faster with large numbers of
training examples and moderate-sized databases.

\section{Experiments}

\begin{table}
\caption{\small Comparison of TensorLog to ProbLog2 and ProPPR} \label{tab:results}
\begin{center}
\begin{tabular}{l|rr}
\hline
          & \multicolumn{2}{c}{Social Influence Task} \\
\hline
ProbLog2  &   20 nodes & 40-50 sec \\
TensorLog & 3327 nodes & 0.84 msec \\
\hline
\end{tabular}\\
~\\
~\\
\begin{tabular}[c]{l|rrr}
\hline
          & \multicolumn{3}{c}{Path-finding} \\
          & \multicolumn{1}{c}{Size} & \multicolumn{1}{c}{Time} & \multicolumn{1}{c}{Acc} \\
\hline
ProbLog2  & 16x16 grid, $d=10$ & 100-120 sec & \\
TensorLog & 16x16 grid, $d=10$ & 5.2 msec  &    \\
(trained) & 16x16 grid, $d=10$ & 18.7 msec  &   96.5\% \\
          & 64x64 grid, $d=64$ & 5.4 msec  &    \\
\hline
\end{tabular}\begin{minipage}[c]{0.3\linewidth}
\includegraphics[width=\linewidth]{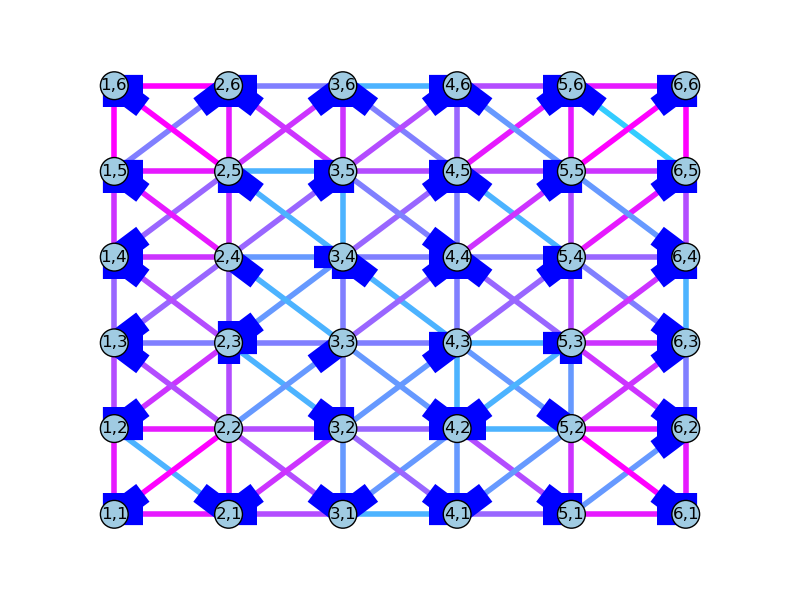}
\end{minipage}\\
~\\
~\\
\begin{tabular}{l|rr|rr} \hline
                      & \multicolumn{2}{c|}{ProPPR} & \multicolumn{2}{c}{TensorLog} \\
                      & \multicolumn{1}{c}{AUC} & \multicolumn{1}{c|}{sec} &
                                                  \multicolumn{1}{c}{AUC} & \multicolumn{1}{c}{sec} \\
\hline
Cora (13k facts,10 rules)
                      & 83.2     & 97.9s    & 97.6   & 102.8s \\
\hline
Wordnet (276k facts)  &          &          &        &          \\
~~Hypernym (46 rules) & 93.4     & 166.8s   & 93.3   & 154.9s   \\
~~Hyponym (46 rules)  & 92.1     & 165.6s   & 92.8   & 152.5s   \\
~~Deriv.~Related (49 rules)  
                      &  8.2     & 166.6s    & 6.7    & 168.2s  \\
\hline
Freebase15k (923k facts)   &          &          &        &     \\
~~division-2nd-level  &  56.4    & 128.5s   &  50.8  & 95.7s    \\
~~person-profession   &  45.8    & 24.4s    &  50.0  & 13.7s    \\
~~actor-performance   &  37.4    & 19.0s    &  38.0  & 13.7s    \\
\hline
\end{tabular}
\end{center}
\end{table}


We compared TensorLog's inference time with ProbLog2, a mature
probabilistic logic programming system which implements the tuple
independence semantics, on two inference problems described in
\cite{fierens2016}.  One is a version of the ``friends and smokers''
problem, a toy model of social influence.  In \cite{fierens2016} small
graphs were artificially generated using a preferential attachment
model, the details of which were not described; instead we used a
small existing network dataset\footnote{The Citeseer dataset from
  \cite{DBLP:conf/asunam/LinC10}.} which displays
preferential-attachment statistics.  The inference times we report are
for the same inference tasks, for a subset of 120 randomly-selected
entities.  In spite of querying six times as many entities, TensorLog
is many times faster.  

We also compare on a path-finding task, also described in
\cite{fierens2016}, which is intended to test performance on deeply
recursive tasks.  The goal here is to compute fixed-depth transitive
closure on a grid: in \cite{fierens2016} a 16-by-16 grid was used,
with a maximum path length of 10.  Again TensorLog shows much faster
performance, and better scalability\footnote{We set TensorLog's
  maximum depth to 10 for the 16-by-16 grid, and to 99 for the larger
  grid.}, as shown by run times on a larger 64-by-64 grid.

These results demonstrate that TensorLog's approximation to ProbLog2's
semantics is efficient, but not that it is useful.  To demonstrate
that TensorLog can efficiently and usefully approximate deeply
recursive concepts, we posed a learning task on the 16-by-16 grid, and
trained TensorLog to approximate the distribution for this task.  The
dataset consists of 256 grid cells connected by 2116 edges, so there
are 256 example queries of the form \cd{path(a,X)} where $a$ is a
particular grid cell.  We picked 1/3 of these queries as test, and the
remainder as train, and trained so that that the single positive
answer to the query \cd{path(a,X)} is the extreme corner closest to
\cd{a}---i.e., one of the corners (1,1), (1,16), (16,1) or (16,16).
Training for 20 epochs brings the accuracy from to 0\% to 96.5\% (for
test), and learning takes approximately 3 sec/epoch.  After learning
query times are still quite fast.  

We note, however, that ProbLog2, in addition to implementing the full
tuple-independence semantics, implements a much more expressive logic
than considered here, including a large portion of full Prolog. In
contrast TensorLog includes only a subset of Datalog.

The table also includes a visualization of the learned weights for a
small 6x6 grid.  For every pair of adjacent grid cells $u,v$, there
are two weights to learn, one for the edge from $u$ to $v$ and one for
its converse.  For each weight pair, we show a single directed edge
(the heavy blue squares are the arrows) colored by the magnitude of
the difference.

We also compared experimentally with ProPPR on several tasks.  One was
a citation-matching task (from \cite{wang2013programming}), in which
ProPPR was favorable compared to MLNs\footnote{We replicated the
  experiments with the most recent version of ProPPR, obtaining a
  result slightly higher than the 2013 version's published AUC of
  80.0}.  Motivated by recent comparisons between ProPPR and
embedding-based approaches to knowledge-base completion
\cite{Wang-Cohen:2016:IJCAI}, we also compared to ProPPR on six
relation-prediction tasks\footnote{We chose this protocol since the
  current TensorLog implementation can only learn parameters for one
  target relation at a time.} involving two databases, Wordnet and
FreeBase15k, a 15,000-entity subset of FreeBase, using rules from the
(non-recursive) theory used in \cite{Wang-Cohen:2016:IJCAI}.

In all of these tasks parameters are learned on a separate training
set.  For TensorLog's learner, we optimized unregularized
cross-entropy loss, using a fixed-rate gradient descent learner.  We
set the learning rate to 0.1, used no regularization, and used a fixed
number of epochs (30), which approximately matched ProPPR's learning
time.\footnote{Since the current TensorLog implementation is
  single-threaded we used only one thread for ProPPR as well.}  The
parameters $\theta_f$ are simply ``clipped'' to prevent them becoming
negative (as in a rectified linear unit) and we use softmax to convert
the output of the $g^p$ functions to distributions.  We used the
default parameters for ProPPR's learning.

Encouragingly, the accuracy of the two systems after learning is
comparable, even with TensorLog's rather simplistic learning scheme.
ProPPR, of course, is not well suited to tight integration with deep
learners.

\section{Concluding Remarks}

Large knowledge bases (KBs) are useful in many tasks, but integrating
this knowledge into deep learners is a challenge.  To address this
problem, we described a probabilistic deductive database, called
TensorLog, in which reasoning is performed with a differentiable
process.  The current TensorLog prototype is limited in many respects:
for instance, it is not multithreaded, and only the simplest learning
algorithms have been tested.  In spite of this, it appears to be
comparable to more mature first-order probabilistic learners in
learning performance and inference time---while holding the promise of
allowing large KBs to be tightly integrated with deep learning.

\subsection*{Acknowledgements}

Thanks to William Wang for providing some of the datasets used here;
and to William Wang, Katie Mazaitis, and many other colleagues
contributed with technical discussions and advice.  The author is
greatful to Google for financial support, and also to NSF for their
support of his work via grants CCF-1414030 and IIS-1250956.

\newpage

\bibliographystyle{plain}
\bibliography{./all}  

\begin{thebibliography}{10}

\bibitem{DBLP:journals/corr/AndreasRDK16}
Jacob Andreas, Marcus Rohrbach, Trevor Darrell, and Dan Klein.
\newblock Learning to compose neural networks for question answering.
\newblock {\em CoRR}, abs/1601.01705, 2016.

\bibitem{brocheler2012probabilistic}
Matthias Brocheler, Lilyana Mihalkova, and Lise Getoor.
\newblock Probabilistic similarity logic.
\newblock In {\em Proceedings of the Conference on Uncertainty in Artificial
  Intelligence}, 2010.

\bibitem{DBLP:journals/ml/Cussens01}
James Cussens.
\newblock Parameter estimation in stochastic logic programs.
\newblock {\em Machine Learning}, 44(3):245--271, 2001.

\bibitem{de2008probabilistic}
Luc De~Raedt and Kristian Kersting.
\newblock {\em Probabilistic inductive logic programming}.
\newblock Springer, 2008.

\bibitem{fierens2016}
Daan Fierens, Guy Van~Den Broeck, Joris Renkens, Dimitar Shterionov, Bernd
  Gutmann, Ingo Thon, Gerda Janssens, and Luc~De Raedt.
\newblock Inference and learning in probabilistic logic programs using weighted
  boolean formulas.
\newblock To appear in Theory and Practice of Logic Programming, 2016.

\bibitem{gormley_approximation-aware_2015}
Matthew~R. Gormley, Mark Dredze, and Jason Eisner.
\newblock Approximation-aware dependency parsing by belief propagation.
\newblock {\em Transactions of the Association for Computational Linguistics
  (TACL)}, 2015.

\bibitem{kimmig2015lifted}
Angelika Kimmig, Lilyana Mihalkova, and Lise Getoor.
\newblock Lifted graphical models: a survey.
\newblock {\em Machine Learning}, 99(1):1--45, 2015.

\bibitem{DBLP:conf/asunam/LinC10}
Frank Lin and William~W. Cohen.
\newblock Semi-supervised classification of network data using very few labels.
\newblock In Nasrullah Memon and Reda Alhajj, editors, {\em ASONAM}, pages
  192--199. IEEE Computer Society, 2010.

\bibitem{poole1997independent}
David Poole.
\newblock The independent choice logic for modelling multiple agents under
  uncertainty.
\newblock {\em Artificial intelligence}, 94(1):7--56, 1997.

\bibitem{RichardsonMLJ2006}
Matthew Richardson and Pedro Domingos.
\newblock Markov logic networks.
\newblock {\em Mach. Learn.}, 62(1-2):107--136, 2006.

\bibitem{rocktaschel2016learning}
Tim Rockt{\"a}schel and Sebastian Riedel.
\newblock Learning knowledge base inference with neural theorem provers.
\newblock In {\em NAACL Workshop on Automated Knowledge Base Construction
  (AKBC)}, 2016.

\bibitem{riedelinjecting2015}
Tim Rocktäschel, Sameer Singh, and Sebastian Riedel.
\newblock Injecting logical background knowledge into embeddings for relation
  extraction.
\newblock In {\em Proc. of ACL/HLT}, 2015.

\bibitem{DBLP:journals/corr/SourekAZK15}
Gustav Sourek, Vojtech Aschenbrenner, Filip Zelezn{\'{y}}, and Ondrej Kuzelka.
\newblock Lifted relational neural networks.
\newblock {\em CoRR}, abs/1508.05128, 2015.

\bibitem{suciu2011probabilistic}
Dan Suciu, Dan Olteanu, Christopher R{\'e}, and Christoph Koch.
\newblock Probabilistic databases.
\newblock {\em Synthesis Lectures on Data Management}, 3(2):1--180, 2011.

\bibitem{TowellAAAI90}
Geoffrey Towell, Jude Shavlik, and Michiel Noordewier.
\newblock Refinement of approximate domain theories by knowledge-based
  artificial neural networks.
\newblock In {\em Proceedings of the Eighth National Conference on Artificial
  Intelligence}, Boston, Massachusetts, 1990. MIT Press.

\bibitem{Wang-Cohen:2016:IJCAI}
William~Yang Wang and William~W. Cohen.
\newblock Learning first-order logic embeddings via matrix factorization.
\newblock In {\em Proceedings of the 25th International Joint Conference on
  Artificial Intelligence (IJCAI 2015)}, New York, NY, July 2016. AAAI.

\bibitem{wang2013programming}
William~Yang Wang, Kathryn Mazaitis, and William~W Cohen.
\newblock Programming with personalized {PageRank}: a locally groundable
  first-order probabilistic logic.
\newblock In {\em Proceedings of the 22nd ACM International Conference on
  Conference on Information \& Knowledge Management}, pages 2129--2138. ACM,
  2013.

\bibitem{wang2014structure}
William~Yang Wang, Kathryn Mazaitis, and William~W Cohen.
\newblock Structure learning via parameter learning.
\newblock In {\em Proceedings of the 23rd ACM International Conference on
  Conference on Information and Knowledge Management}, pages 1199--1208. ACM,
  2014.

\end{thebibliography}

\end{document}